%% file: main.tex
\documentclass[10pt,twocolumn,letterpaper]{article}

\usepackage[pagenumbers]{cvpr} 


\usepackage{xcolor}
\usepackage{multirow}
\usepackage{algorithm}
\usepackage{algorithmic}
\usepackage{amsfonts}
\usepackage{graphicx}

\usepackage{bbm}
\usepackage{float}
\usepackage{circledsteps}
\definecolor{cvprblue}{rgb}{0.21,0.49,0.74}
\usepackage[pagebackref,breaklinks,colorlinks,allcolors=cvprblue]{hyperref}
\usepackage[T1]{fontenc}

\def\paperID{5270} 
\def\confName{CVPR}
\def\confYear{2025}
\title{Any3DIS: Class-Agnostic 3D Instance Segmentation by 2D Mask Tracking}

\author{
Phuc Nguyen$^{1}$ \quad Minh Luu$^{1}$ \quad Anh Tran$^{1}$ \quad Cuong Pham$^{1,2}$ \quad Khoi Nguyen$^{1}$ \\
\normalsize{$^1$VinAI Research \quad $^2$Posts \& Telecommunications Inst. of Tech.} \\
\tt\small \{v.phucnda,  v.minhlnh, v.anhtt152, v.khoindm\}@vinai.io \qquad cuongpv@ptit.edu.vn  \\
{\url{https://any3dis.github.io/}}
}

\def\Approach{Any3DIS}

\begin{document}
\input{definition}

\twocolumn[{
\renewcommand\twocolumn[1][]{#1}%
\maketitle
\vspace{-30pt}
\begin{center}%
\includegraphics[width=.95\linewidth]{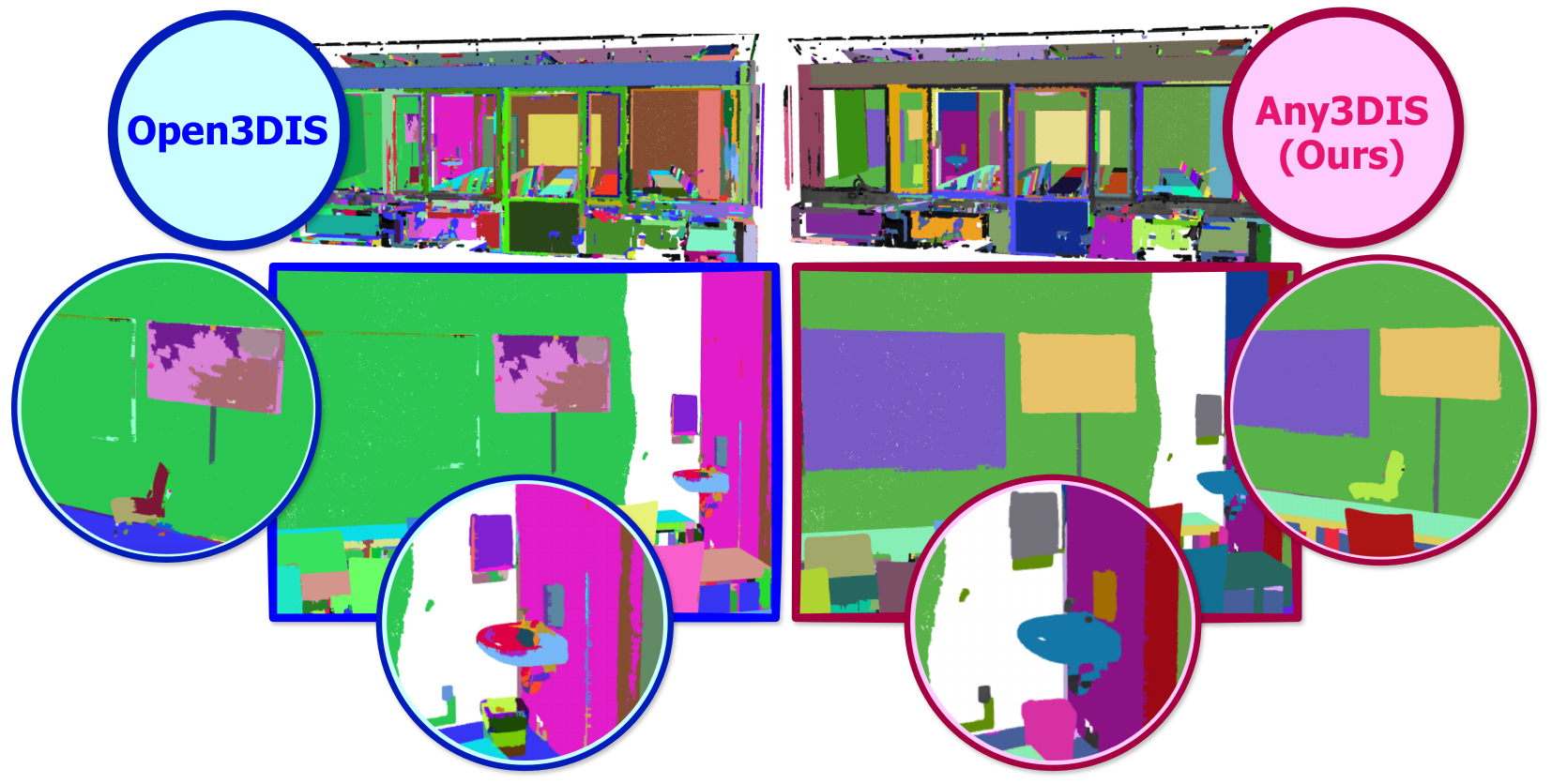}
\vspace{-5pt}
\captionof{figure}{Comparison of our proposed approach, \Approach, with existing 3D instance segmentation methods such as Open3DIS \cite{nguyen2023open3dis}. Open3DIS frequently encounters over-segmentation issues, generating redundant 3D proposals due to its unsupervised merging process. In contrast, our approach leverages robust guidance from 2D mask tracking to maintain consistent object segmentation across video frames, effectively reducing redundant proposals and enhancing segmentation accuracy.
}
\label{fig:teaser}
\end{center}
}]

\input{sec/0_abstract}    
\input{sec/1_intro}
\input{sec/2_relatedwork}

\input{sec/3_method}

\input{sec/4_experiments}

\input{sec/5_conclusion}

{
    \small
    \bibliographystyle{ieeenat_fullname}
    \bibliography{main}
}


\end{document}

%% file: definition.tex
\def\mA{\mathcal{A}}
\def\mB{\mathcal{B}}
\def\mC{\mathcal{C}}
\def\mD{\mathcal{D}}
\def\mE{\mathcal{E}}
\def\mF{\mathcal{F}}
\def\mG{\mathcal{G}}
\def\mH{\mathcal{H}}
\def\mI{\mathcal{I}}
\def\mJ{\mathcal{J}}
\def\mK{\mathcal{K}}
\def\mL{\mathcal{L}}
\def\mM{\mathcal{M}}
\def\mN{\mathcal{N}}
\def\mO{\mathcal{O}}
\def\mP{\mathcal{P}}
\def\mQ{\mathcal{Q}}
\def\mR{\mathcal{R}}
\def\mS{\mathcal{S}}
\def\mT{\mathcal{T}}
\def\mU{\mathcal{U}}
\def\mV{\mathcal{V}}
\def\mW{\mathcal{W}}
\def\mX{\mathcal{X}}
\def\mY{\mathcal{Y}}
\def\mZ{\mathcal{Z}} 

\def\bbN{\mathbb{N}} 
\def\bbR{\mathbb{R}} 
\def\bbP{\mathbb{P}} 
\def\bbQ{\mathbb{Q}} 
\def\bbE{\mathbb{E}}

\def\1n{\mathbf{1}_n}
\def\0{\mathbf{0}}
\def\1{\mathbf{1}}

\def\A{{\bf A}}
\def\B{{\bf B}}
\def\C{{\bf C}}
\def\D{{\bf D}}
\def\E{{\bf E}}
\def\F{{\bf F}}
\def\G{{\bf G}}
\def\H{{\bf H}}
\def\I{{\bf I}}
\def\J{{\bf J}}
\def\K{{\bf K}}
\def\L{{\bf L}}
\def\M{{\bf M}}
\def\N{{\bf N}}
\def\O{{\bf O}}
\def\P{{\bf P}}
\def\Q{{\bf Q}}
\def\R{{\bf R}}
\def\S{{\bf S}}
\def\T{{\bf T}}
\def\U{{\bf U}}
\def\V{{\bf V}}
\def\W{{\bf W}}
\def\X{{\bf X}}
\def\Y{{\bf Y}}
\def\Z{{\bf Z}}

\def\a{{\bf a}}
\def\b{{\bf b}}
\def\c{{\bf c}}
\def\d{{\bf d}}
\def\e{{\bf e}}
\def\f{{\bf f}}
\def\g{{\bf g}}
\def\h{{\bf h}}
\def\i{{\bf i}}
\def\j{{\bf j}}
\def\k{{\bf k}}
\def\l{{\bf l}}
\def\m{{\bf m}}
\def\n{{\bf n}}
\def\o{{\bf o}}
\def\p{{\bf p}}
\def\q{{\bf q}}
\def\r{{\bf r}}
\def\s{{\bf s}}
\def\t{{\bf t}}
\def\u{{\bf u}}
\def\v{{\bf v}}
\def\w{{\bf w}}
\def\x{{\bf x}}
\def\y{{\bf y}}
\def\z{{\bf z}}

\def\balpha{\mbox{\boldmath{$\alpha$}}}
\def\bbeta{\mbox{\boldmath{$\beta$}}}
\def\bdelta{\mbox{\boldmath{$\delta$}}}
\def\bgamma{\mbox{\boldmath{$\gamma$}}}
\def\blambda{\mbox{\boldmath{$\lambda$}}}
\def\bsigma{\mbox{\boldmath{$\sigma$}}}
\def\btheta{\mbox{\boldmath{$\theta$}}}
\def\bomega{\mbox{\boldmath{$\omega$}}}
\def\bxi{\mbox{\boldmath{$\xi$}}}
\def\bnu{\mbox{\boldmath{$\nu$}}}                                  
\def\bphi{\mbox{\boldmath{$\phi$}}}
\def\bmu{\mbox{\boldmath{$\mu$}}}

\def\bDelta{\mbox{\boldmath{$\Delta$}}}
\def\bOmega{\mbox{\boldmath{$\Omega$}}}
\def\bPhi{\mbox{\boldmath{$\Phi$}}}
\def\bLambda{\mbox{\boldmath{$\Lambda$}}}
\def\bSigma{\mbox{\boldmath{$\Sigma$}}}
\def\bGamma{\mbox{\boldmath{$\Gamma$}}}
                                  
\newcommand{\myprob}[1]{\mathop{\mathbb{P}}_{#1}}

\newcommand{\myexp}[1]{\mathop{\mathbb{E}}_{#1}}

\newcommand{\mydelta}[1]{1_{#1}}

\newcommand{\myminimum}[1]{\mathop{\textrm{minimum}}_{#1}}
\newcommand{\mymaximum}[1]{\mathop{\textrm{maximum}}_{#1}}    
\newcommand{\mymin}[1]{\mathop{\textrm{minimize}}_{#1}}
\newcommand{\mymax}[1]{\mathop{\textrm{maximize}}_{#1}}
\newcommand{\mymins}[1]{\mathop{\textrm{min.}}_{#1}}
\newcommand{\mymaxs}[1]{\mathop{\textrm{max.}}_{#1}}  
\newcommand{\myargmin}[1]{\mathop{\textrm{argmin}}_{#1}} 
\newcommand{\myargmax}[1]{\mathop{\textrm{argmax}}_{#1}} 
\newcommand{\myst}{\textrm{s.t. }}

\newcommand{\denselist}{\itemsep -1pt}
\newcommand{\sparselist}{\itemsep 1pt}

\definecolor{pink}{rgb}{0.9,0.5,0.5}
\definecolor{purple}{rgb}{0.5, 0.4, 0.8}   
\definecolor{gray}{rgb}{0.3, 0.3, 0.3}
\definecolor{mygreen}{rgb}{0.2, 0.6, 0.2}

\newcommand{\cyan}[1]{\textcolor{cyan}{#1}}
\newcommand{\blue}[1]{\textcolor{blue}{#1}}
\newcommand{\magenta}[1]{\textcolor{magenta}{#1}}
\newcommand{\pink}[1]{\textcolor{pink}{#1}}
\newcommand{\green}[1]{\textcolor{green}{#1}} 
\newcommand{\gray}[1]{\textcolor{gray}{#1}}    
\newcommand{\mygreen}[1]{\textcolor{mygreen}{#1}}    
\newcommand{\purple}[1]{\textcolor{purple}{#1}}       

\definecolor{greena}{rgb}{0.4, 0.5, 0.1}
\newcommand{\greena}[1]{\textcolor{greena}{#1}}

\definecolor{bluea}{rgb}{0, 0.4, 0.6}
\newcommand{\bluea}[1]{\textcolor{bluea}{#1}}
\definecolor{reda}{rgb}{0.6, 0.2, 0.1}
\newcommand{\reda}[1]{\textcolor{reda}{#1}}

\def\changemargin#1#2{\list{}{\rightmargin#2\leftmargin#1}\item[]}
\let\endchangemargin=\endlist
                                               
\newcommand{\cm}[1]{}

\newcommand{\mhoai}[1]{{\color{magenta}\textbf{[MH: #1]}}}

\newcommand{\mtodo}[1]{{\color{red}$\blacksquare$\textbf{[TODO: #1]}}}
\newcommand{\myheading}[1]{\vspace{1ex}\noindent \textbf{#1}}
\newcommand{\htimesw}[2]{\mbox{$#1$$\times$$#2$}}


\newif\ifshowsolution
\showsolutiontrue

\ifshowsolution  
\newcommand{\Solution}[2]{\paragraph{\bf $\bigstar $ SOLUTION:} {\sf #2} }
\newcommand{\Mistake}[2]{\paragraph{\bf $\blacksquare$ COMMON MISTAKE #1:} {\sf #2} \bigskip}
\else
\newcommand{\Solution}[2]{\vspace{#1}}
\fi

\newcommand{\truefalse}{
\begin{enumerate}
	\item True
	\item False
\end{enumerate}
}

\newcommand{\yesno}{
\begin{enumerate}
	\item Yes
	\item No
\end{enumerate}
}

\newcommand{\Sref}[1]{Sec.~\ref{#1}}
\newcommand{\Eref}[1]{Eq.~(\ref{#1})}
\newcommand{\Fref}[1]{Fig.~\ref{#1}}
\newcommand{\Tref}[1]{Table~\ref{#1}}

\definecolor{gray}{rgb}{0.3, 0.3, 0.3}

%% file: sec/0_abstract.tex
\begin{abstract}
\vspace{-25pt}

Existing 3D instance segmentation methods frequently encounter issues with over-segmentation, leading to redundant and inaccurate 3D proposals that complicate downstream tasks. This challenge arises from their unsupervised merging approach, where dense 2D instance masks are lifted across frames into point clouds to form 3D candidate proposals without direct supervision. These candidates are then hierarchically merged based on heuristic criteria, often resulting in numerous redundant segments that fail to combine into precise 3D proposals. To overcome these limitations, we propose a 3D-Aware 2D Mask Tracking module that uses robust 3D priors from a 2D mask segmentation and tracking foundation model (SAM-2) to ensure consistent object masks across video frames. Rather than merging all visible superpoints across views to create a 3D mask, our 3D Mask Optimization module leverages a dynamic programming algorithm to select an optimal set of views, refining the superpoints to produce a final 3D proposal for each object. Our approach achieves comprehensive object coverage within the scene while reducing unnecessary proposals, which could otherwise impair downstream applications. Evaluations on ScanNet200 and ScanNet++ confirm the effectiveness of our method, with improvements across Class-Agnostic, Open-Vocabulary, and Open-Ended 3D Instance Segmentation tasks.

\end{abstract}

%% file: sec/1_intro.tex
\section{Introduction}

\begin{figure}[t]
  \centering
  \includegraphics[width=0.98\linewidth]{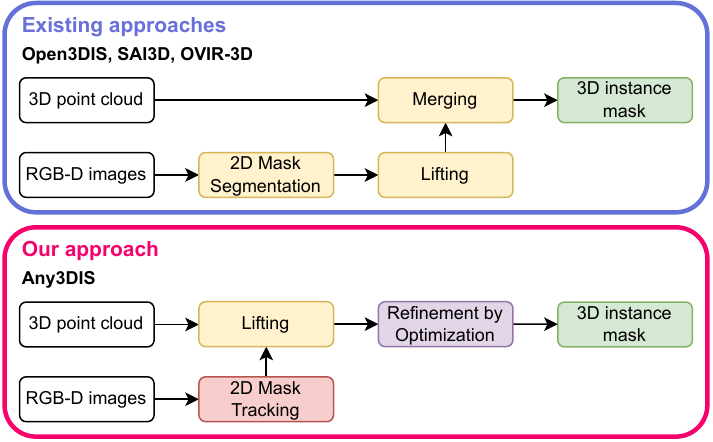}
   \vspace{-5pt}
   \caption{\textbf{Difference between Open3DIS \cite{nguyen2023open3dis} and our \Approach}. Open3DIS segments all 2D masks across views first, then lifts them to 3D proposals for merging, whereas our approach tracks all 2D masks across views, lifts them directly to 3D proposals, and then refines these proposals through optimization.}
   \label{fig:approach}
   \vspace{-14pt}
\end{figure}

3D Instance Segmentation (3DIS) is a core task in computer vision and robotics, with applications spanning autonomous navigation \cite{behley2019iccv}, augmented reality \cite{dehghan2021arkitscenes}, and scene understanding \cite{dai2017scannet}. \textbf{Fully supervised} 3DIS approaches \cite{vu2022softgroup, schult2023mask3d, ngo2023isbnet} rely on extensive manual annotation of 3D point clouds for each semantic class. Although these methods deliver high-quality segmentation results in a \textit{closed-set setting} \cite{straub2019replica, yeshwanthliu2023scannetpp}, they struggle to generalize to previously unseen objects.

Recent research has explored class-agnostic 3D instance segmentation (3DIS) to overcome this limitation. \textbf{Training-based} approaches \cite{huang2025segment3d, guo2024sam-graph} leverage pseudo labels generated by 2D foundation models such as SAM \cite{kirillov2023segment} to train a 3D instance segmenter. These methods use point-pixel mapping to lift 2D masks directly into 3D pseudo masks. However, they are constrained by the quality of the pseudo labels, which are usually non-reliable.
Alternatively, \textbf{training-free} or \textbf{unsupervised-merging during testing} methods \cite{lu2023ovir, yang2023sam3d, yin2023sai3d, xu2023sampro3d, nguyen2023open3dis} tackle this issue by lifting 2D segmentation masks to 3D space without additional training and merging the lifted masks using heuristic criteria to generate 3D proposals for novel objects. While promising, these methods encounter notable challenges. The 3D scene segmentation is prone to over-segmentation due to dense and inconsistent mask predictions across multiple views as shown in \cref{fig:teaser}. Additionally, the computational cost is high, as the lack of guided merging criteria necessitates retaining all possible mask combinations for safety.

To address these limitations, we introduce a novel approach termed \textbf{3D Instance Segmentation by 2D Mask Tracking} which is compared to the previous method Open3DIS \cite{nguyen2023open3dis} in \cref{fig:approach}. This method leverages the 2D segmentation and tracking capabilities of the foundation model SAM 2 \cite{ravi2024sam}, generating comprehensive 2D mask tracks from an input RGB-D sequence. Each track can then be transformed into a potential 3D proposal, simplifying the complex merging process to a straightforward task of 2D mask tracking. However, directly applying SAM 2 on RGB frames alone is suboptimal, as it disregards valuable 3D information from the accompanying point cloud. To overcome this, we propose \textbf{3D-Aware 2D Mask Tracking}, which integrates geometric cues from the 3D point cloud to enhance the robustness and accuracy of 2D mask tracking.

Next, we lift the 2D mask tracks to 3D by aggregating superpoints whose 2D projections overlap with any of the track’s 2D masks above a given threshold. However, directly selecting all such 3D superpoints may lead to suboptimal results, as it overlooks each superpoint’s multi-view consistency. Specifically, while a superpoint might substantially overlap with a 2D mask in one view, it may have minimal or no overlap in other views. To address this, we propose \textbf{3D Proposal Refinement via Optimization} approach, selectively assigning superpoints to a 3D proposal based on their cumulative overlap across all views. Although this optimization is NP-hard, we present a dynamic programming algorithm to obtain a fast and effective solution.

We evaluate our method on both the ScanNet200 \cite{rozenberszki2022language} and ScanNet++ \cite{yeshwanthliu2023scannetpp} datasets. 
Our approach demonstrates substantial improvements in class-agnostic segmentation, outperforming Open3DIS and other fully supervised models on both datasets. Furthermore, our method achieves enhanced results on Open-Vocabulary \cite{takmaz2023openmask3d, nguyen2023open3dis} and Open-Ended \cite{nguyen2024open} 3D Instance Segmentation tasks, establishing a scalable solution adaptable to diverse, real-world applications and capable of handling various object classes and scene complexities.

Our contributions can be summarized as follows:
\begin{itemize}
    \item We propose a \textbf{3D-Aware 2D Mask Tracking} module that leverages strong 3D cues from a 2D foundation mask-tracking model, effectively producing consistent object-level masks across video frames using SAM-2.
    \item We introduce a \textbf{3D Proposal Refinement via Optimization} process that employs dynamic programming to generate high-quality 3D proposals, significantly minimizing redundant 3D proposals.
    \item We validate the effectiveness of our method on the ScanNet200 \cite{dai2017scannet} and ScanNet++ \cite{yeshwanthliu2023scannetpp} datasets, achieving SOTA performance on Class-Agnostic, Open-Vocabulary, and Open-Ended 3D Instance Segmentation tasks.
\end{itemize}

\label{sec:intro}

%% file: sec/2_relatedwork.tex
\section{Related Work}
\label{sec:relatedwork}

\myheading{Fully-Supervised 3D Instance Segmentation} methods, such as SoftGroup \cite{vu2022softgroup}, Mask3D \cite{schult2023mask3d}, and ISBNet \cite{ngo2023isbnet}, trained on closed-set datasets \cite{straub2019replica, yeshwanthliu2023scannetpp, dehghan2021arkitscenes, rozenberszki2022language}, are well-regarded for delivering reliable 3D segmentation results. These approaches use 3D convolutional neural networks \cite{mink, qi2017pointnet, qi2017pointnet++} to capture detailed semantic information from 3D point cloud scenes, employing clustering \cite{dbscan, jiang2020pointgroup} or kernel prediction \cite{He2021dyco3d, ngo2023isbnet} techniques to generate final 3D instance segmentations. While these methods achieve high-quality, fine-grained 3D proposals, they have a critical limitation: they cannot detect or localize objects beyond the training set, which restricts their utility in open-world environments.

\myheading{Class-Agnostic 3D Instance Segmentation} tries to segment all possible 3D object masks regardless of their semantic labels.
The approaches can be broadly classified into two categories: training-based and training-free methods.
\textbf{Training-based} methods leverage 3D annotations and pseudo labels from 2D foundation models to train an instance decoder. For example, Segment3D \cite{huang2025segment3d} uses pseudo labels generated by SAM \cite{kirillov2023segment} in conjunction with ground-truth masks to guide a two-stage training process with Minkowski’s Unet \cite{mink} and a MaskFormer \cite{cheng2021maskformer} decoder. Similarly, SAM-Graph \cite{guo2024sam-graph} initializes its Graph Neural Network with SAM \cite{kirillov2023segment} and CropFormer \cite{qilu2023high} pseudo labels. However, these methods are fine-tuned using ground-truth masks, which limits their ability to segment unseen object classes.
\textbf{Training-free} approaches seek to overcome this limitation by directly generating 3D proposals from 2D segmentation masks produced by 2D foundation models. Methods like OVIR-3D \cite{lu2023ovir}, SAM3D \cite{yang2023sam3d}, SAI3D \cite{yin2023sai3d}, SAMPro3D \cite{xu2023sampro3d}, and Open3DIS \cite{nguyen2023open3dis} utilize models such as SAM to produce dense mask predictions for \textit{every} 2D RGB-D frame, which are then aggregated into a unified set of 3D proposals. While these methods improve the quality of mask predictions, 
inconsistencies in 2D segmentation across RGB-D frames hinder their robustness, as they rely on heuristic merging criteria.

\myheading{Open-Vocabulary 3D Instance Segmentation (OV-3DIS)} focuses on segmenting 3D objects from classes introduced during testing, rather than those seen during training. Methods like OpenMask3D~\cite{takmaz2023openmask3d} and Lowis3D~\cite{ding2024lowis3d} use 3D instance segmentation (3DIS) networks~\cite{ngo2023isbnet,schult2023mask3d} to produce class-agnostic proposals. In contrast, approaches such as SAI3D~\cite{yin2023sai3d}, MaskClustering~\cite{yan2024maskclustering}, and OVIR-3D~\cite{lu2023ovir} employ 2D segmenters to generate masks for each view, which are then lifted into 3D space.
Each method has its strengths and limitations: 3DIS networks excel at capturing large geometric structures but may miss small or rare objects due to class imbalance, while 2D segmenters are better at detecting small regions but struggle to maintain object consistency when projecting their results into 3D. Open3DIS~\cite{nguyen2023open3dis} combines both 3D and 2D segmentation branches, leveraging the strengths of each to capture small objects while preserving larger structures through superpoint-level masks. However, OV-3DIS still relies on a predefined vocabulary during testing, requiring human intervention and limiting its use in truly autonomous systems that must operate without prior knowledge of object classes.

\myheading{Open-Ended 3D Instance Segmentation (OE-3DIS)} was introduced by prior work \cite{nguyen2024open} to enable object recognition without relying on predefined class labels during training or testing. This method uses Multimodal Large Language Models (MLLMs) to generate features in 2D space, which are then projected into the 3D point cloud. The results show that OE-3DIS can outperform Open3DIS, the current state-of-the-art in OV-3DIS, on the ScanNet++ dataset, without requiring ground-truth class names.

%% file: sec/3_method.tex
\section{Any3DIS}
\label{sec:method}

\begin{figure*}[t]
  \centering
  \includegraphics[width=.97\linewidth]{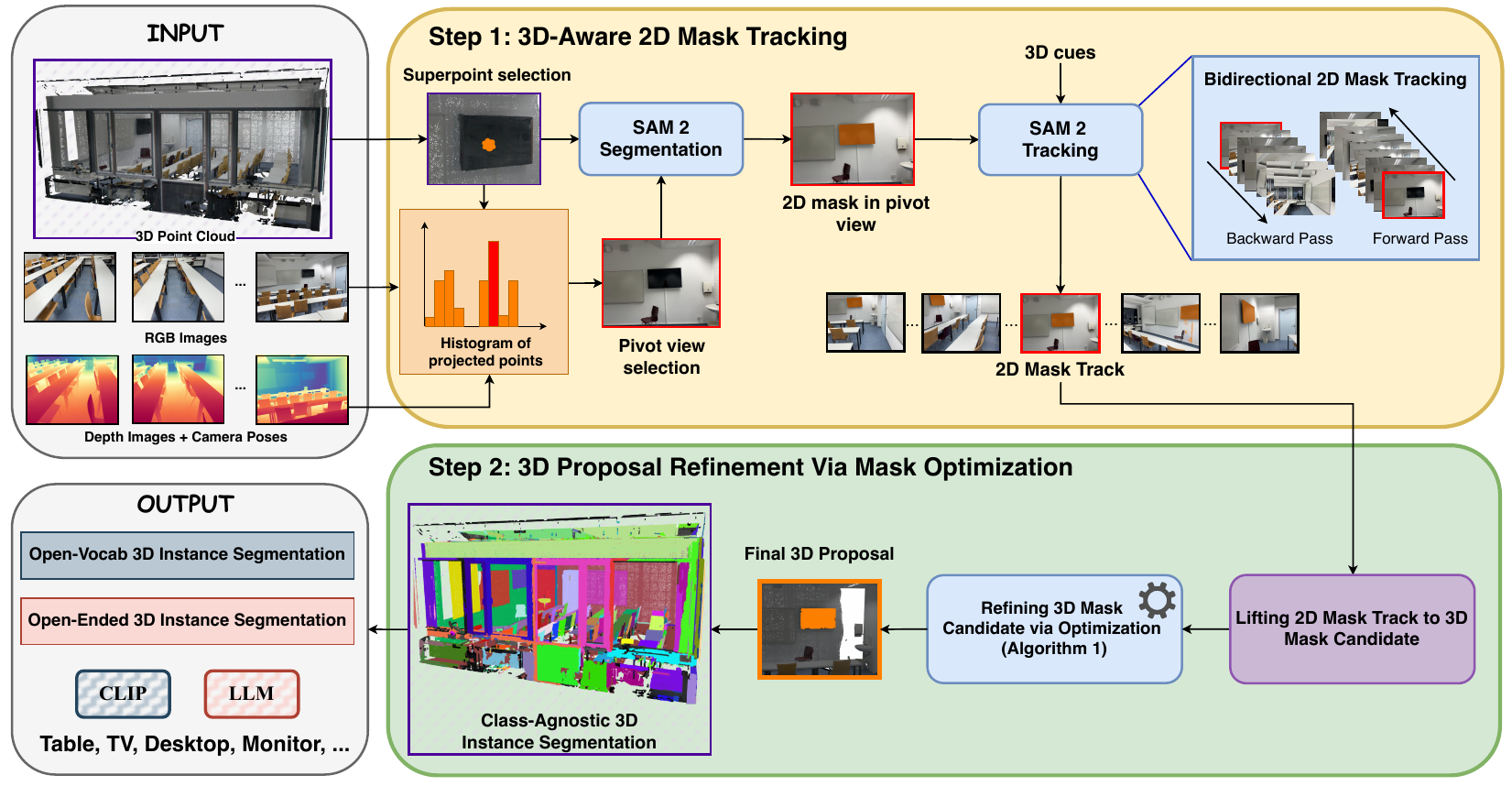}
   \vspace{-7pt}
   \caption{\textbf{Overview of \Approach}. We propose a novel class-agnostic approach for 3D instance segmentation through 2D mask tracking. Specifically, we identify each selected superpoint's ``pivot view'' within the RGB-D frame sequence, where it is most visible. Using the SAM 2 model \cite{ravi2024sam}, we then obtain the 2D segmentation of this superpoint in the pivot view and track this 2D segmentation across other views. For each 2D mask obtained, we generate a 3D mask candidate by aggregating all 3D superpoints whose projections intersect with any of the 2D masks. These superpoints are then subjected to mask optimization to produce the final refined 3D mask proposal, which is added to the class-agnostic 3D instance segmentation bank, ready for downstream tasks.
   }
   \vspace{-12pt}
   \label{fig:architecture}
\end{figure*}

\textbf{Problem Definition:} Given a 3D point cloud scene $\{\mathbf{P}_i\}_{i=1}^{N} \in \mathbb{R}^{N \times 6}$ with $N$ points, each point is defined by its spatial coordinates and color values $(x, y, z, r, g, b)$. Additionally, we have $T$ RGB-D frames with the shape of $(H,W)$, each consisting of a color image $\{\mathbf{I}_t\}_{t=1}^T$ and a depth image $\{\mathbf{D}_t\}_{t=1}^T$, where $\mathbf{I}_t \in \mathbb{R}^{H \times W \times 3}$ and $\mathbf{D}_t \in \mathbb{R}_+^{H \times W}$. Every frame $t$ also has the extrinsic matrix $\E_t$ and shares the same intrinsic matrix $\K$ to support the calculation of corresponding pixel projections from the point cloud. Our goal is to segment all $Q$ 3D object binary masks, represented as $\{\mathbf{M}_q\}_{q=1}^{Q}$, where each mask is a binary mask $\mathbf{M}_q \in \{0, 1\}^N$. To facilitate the segmentation as in \cite{nguyen2023open3dis, ngo2023isbnet, huang2025segment3d, yin2023sai3d}, we further perform a normal-based graph cut algorithm \cite{felzenszwalb2004efficient} to partition the point cloud into a set of $L$ superpoints $\{\mathbf{S}_l\}_{l=1}^{L}$, where each $\S_l \in \{0, 1\}^N$.

\myheading{Overview:} An overview of \Approach~ is shown in \cref{fig:architecture}. Our approach leverages the concept of \textbf{3D instance segmentation by 2D mask tracking}. First, in \cref{sec:3d-aware-tracking}, we aim to find the tracks of all possible objects across frames by sampling superpoints. For each sampled superpoint, the 3D-Aware Instance Tracking module is applied to track consistent 2D segmentation masks across $T$ frames for the corresponding object. In \cref{sec:mask-opt}, these tracks are then lifted into 3D mask candidates by selecting superpoints whose projections sufficiently overlap with the 2D masks of these tracks. Finally, the 3D candidates are refined using the Mask Optimization module to retain the most relevant superpoints for the final 3D proposals.

\subsection{3D-Aware 2D Mask Tracking}
\label{sec:3d-aware-tracking}

The goal of this step is to track all possible objects across the RGB-D sequence. However, obtaining complete tracks for each object is challenging, as objects vary in appearance across frames in both spatial and temporal dimensions. For instance, if we begin tracking all visible objects in the first frame and continue until they disappear, we risk missing objects that appear in later frames. Additionally, the object’s initial mask in the first frame does not often represent its largest visible area, which can reduce tracking quality. 

To address these challenges, we propose a novel approach to retrieve comprehensive tracks of all objects in the 3D scene by leveraging precomputed superpoints $\{\mathbf{S}_l\}_{l=1}^{L}$ as 3D cues representing potential 3D proposals. 
To begin, we use the Farthest Point Sampling (FPS) \cite{eldar1997farthest} technique to select $Q_1$ initial superpoints ($Q_1 \ll L$).
For each selected superpoint, we estimate the most suitable view called \textbf{pivot view} to initiate tracking for the corresponding 3D object, using its projected points as the query to obtain the 2D segmentation in that pivot view and tracking in other views.

\myheading{Pivot view selection.}
For each sampled superpoint $\mathbf{S}_l$, we generate a track of masks for the corresponding object across a sequence of $T$ frames. First, we calculate the \textbf{histogram of projected points} for the superpoint $\mathbf{S}_l$ across these $T$ views, yielding $\{\psi_t^l\}_{t=1}^T$, i.e., counting the number of projected points that can be seen in each view $t$. 

Particularly, for superpoint $\mathbf{S}_l$, its 2D set of projected points $\rho^l_t$ on frame $t$ is obtained by projecting function $\Pi(\cdot)$ using the camera parameters $\K, \E_t$ and depth $\D_t$ as:
\begin{equation}
    \label{equa:pimapper}
    {\rho}^l_t=\Pi(\S_l, \K, \E_t,\mathbf{D}_t). 
\end{equation}
Furthermore, for a 3D superpoint $\mathbf{S}_l$ in a view $t$, its neighboring superpoints (retrieved by the $\operatorname{KNN}$ algorithm using Euclidean distances between centroids) provide valuable context information to help decide whether that view is suitable for the pivot view for tracking. In other words, if any of the $\kappa$ neighboring superpoints $\mathbf{S}_k$ of $\mathbf{S}_l$ are invisible in the current view, signified by low $|\rho_t^k| / N_{\mathbf{S}_k}$, where $N_{\mathbf{S}_k}$ and $|\rho_t^k|$ are the number of 3D points of superpoint $\mathbf{S}_k$ and its number of projected points in view $t$, this will indicate that the view is not suitable even if most of the points of $\mathbf{S}_l$ are visible, and vice versa. This motivates us to multiply the number of projected points of the superpoint in that view by a scale factor $s_t^l \in [0, 1]$, computed as:
\begin{gather}
    \label{equa:weighted}
    s^l_t = \frac{\sum_{k \in \operatorname{KNN}(\mathbf{S}_l)}{\left|{\rho}^k_t\right|/N_{\mathbf{S}_k}}}{\kappa}.
\end{gather}
The final histogram of projected points of superpoint $\S_l$ in view $t$ is:
$
    \psi_t^l = |\rho^l_t| \cdot s_t^l.
$
Next, we pick the view with the highest value from the histogram as the `\textit{pivot view}’ for superpoint $\mathbf{S}_l$, denoted as $\text{pivot}(l) = \operatorname{argmax}_t \psi_t^l$.

\myheading{Segmentation in pivot view and tracking in other views.}
After choosing the pivot view, we simply sample three projected points from $\rho^l_{\text{pivot}(l)}$ using FPS as the \textit{point prompting queries} for SAM 2 \cite{ravi2024sam}, a 2D foundation model for segmentation and tracking, to leverage its powerful tracking capability. We obtain the 2D mask with \textbf{score} $o_l$ in the pivot frame and use it as the \textit{mask prompting query} for tracking in other views. Specifically,
we construct two tracking sequences: one backward $[\mathbf{I}_{\text{pivot}(l)}:\mathbf{I}_{0}]$ and one forward $[\mathbf{I}_{\text{pivot}(l)}:\mathbf{I}_{T}]$. It is worth noting that an object can disappear and later reappear in subsequent frames. SAM 2 can track these reappearing objects within a predefined memory window of 7 frames; after this, SAM 2 forgets them and treats them as new objects. To address this, we sample a projected point of superpoint $\mathbf{S}_l$ at the reappearance view $t$ from $\rho^l_t$ as an \textit{additional point query} for SAM 2, enabling consistent tracking of the target object over a longer sequence.

At the end of this step, we obtain $Q_1$ 2D binary mask tracks $\{\m_q^t\}_{t\in [1..T], q\in [1..Q_1]}$ with scores $\{o_q\}_{q\in [1..Q_1]}$ from the RGB-D sequence, representing $Q_1$ possible 3D objects. 

\subsection{3D Proposal Refinement via Optimization}
\label{sec:mask-opt}
The goal of this step is to generate a final 3D mask proposal for each 2D mask track composed of 3D superpoints.

\myheading{Lifting 2D mask track to 3D mask candidate.} For each 2D mask track $\mathbf{m}_q$ of 3D object $q$, we lift it to 3D by grouping all superpoints $\mathbf{S}_l$ whose projections overlap with any 2D mask $\mathbf{m}_q^t$ in any view $t$ by more than a threshold $\tau$. We first compute the binary value $\overline{\M}_q^{t,l}$ indicating the visibility of superpoint $\S_l$ in view $t$  as follows:
\begin{gather}
    \overline{\M}_q^{t,l} =  
    \begin{cases} 
        1, & \text{if } \operatorname{IOU}(\rho^l_t, \m_q^t) \geq \tau  \\ 
        0, & \text{otherwise} 
    \end{cases}
    \text{ for } l \in [1..L]. \label{eq:visible_set}
\end{gather}
Note that $\overline{\M}_q^t \in \{0, 1\}^L$ is binary vector over all superpoints and $\overline{\M}_q$ is the 3D mask candidate of object $q$ by merging all visible superpoints in $\overline{\M}_q^t$ over all views.
Nevertheless, selecting all these superpoints may lead to suboptimal outcomes, as it overlooks the multi-view consistency of each superpoint. Notably, while a superpoint $\S_l$ might significantly overlap with a 2D mask in one view, it could have little to no overlap in other views. To tackle this issue, we further \textbf{refine} all superpoints $\S_l$ of 3D mask candidate $\overline{\M}_q$ to form the final 3D mask proposal $\M_q$ which maximizes the alignment with all 2D masks $\m_q^t$ in all views. 

\myheading{Refining 3D mask candidate to final 3D mask proposal.}
To this end, we introduce an \textit{unconstrained binary optimization problem} to decide the membership $\theta=\{0, 1\}^L$ of all superpoints visible in $\S_l \in \overline{\M}_q$ in the final 3D mask proposal $\M_q$ as follows:
\begin{equation}
    \max_\theta \mL(\theta) = \max_\theta \sum_t\underbrace{\rho^\theta_t \otimes \m^q_t}_{\text{(A) Inside 2D Mask}}- \underbrace{\rho^\theta_t \otimes (\1_{H\times W} - \m^q_t)}_{\text{(B) Outside 2D Mask}},
    \label{eq:objective}
\end{equation}
where $\rho^\theta_t$ represents the projection of the current solution $\theta$ in view $t$, computed by projecting the 3D mask $\M_\theta$ formed by merging all superpoints $\S_l$ where $\theta^l = 1$ (i.e., $\M_\theta = \left\{\S_l | \theta^l = 1 \right\}$) onto view $t$. Additionally, $a \otimes b$ counts the number of points in $a$ that belong to $b$. The objective is to maximize the number of points from the current solution $\theta$ that lie within a mask (part (A) in \cref{eq:objective}) and minimize the number of points that lie outside the mask (part (B) in \cref{eq:objective}) across all views. In other words, the 3D mask $\M_\theta$ formed by $\theta$ should align well with the 2D masks in every view to form the optimal final mask $\M_q$.

Solving the binary optimization problem in \cref{eq:objective} is NP-hard since we need to try all possible configurations of $\theta$ to find the optimal solution, which cannot run in polynomial time. Instead, we propose a simple, yet effective dynamic programming algorithm to obtain a fast and efficient solution.

Concretely, for each 2D mask track $\m_q$ of the $q$-th 3D object, we pre-compute its binary vector $\overline{\M}^{t}_q$ as in \cref{eq:visible_set} indicating the visibility of all superpoints in view $t$. Let $\theta_t \in \{0, 1\}^L$ be the current solution up to view $t$ and $\C_t \in \mathbb{N}$ be the objective value $\mL(\theta_t)$  of the current solution $\theta_t$ from \cref{eq:objective}.
We then traverse all views, and at each view $t$, we have two choices: (1) retain the current solution, or (2) add all superpoints in $\overline{\M}^t_q$ to the current solution. We select the option that yields the better objective value $\mL(\theta_t)$ and proceed to the next view. The complete algorithm is outlined in \cref{alg:optimize}. Note that at each view, we choose to either add all newly visible superpoints or none, rather than selectively adding individual superpoints, partial inclusion of superpoints is not considered. Thus, this approach may not guarantee a globally optimal solution, however, it produces high-quality solutions that outperform previous methods in practice as shown in experiments.

\myheading{Iterative 3D object sampling.}
At the end of this stage, we obtain an initial set of $Q_1$ binary masks $\{\M_q\}_{q=1}^{Q_1}$ with corresponding 2D mask scores $\{o_q\}_{q=1}^{Q_1}$ for evaluation. Some superpoints, however, remain unassigned to any 3D mask and may represent new 3D objects. To capture these additional objects, we repeat the process in \cref{sec:3d-aware-tracking} and \cref{sec:mask-opt} iteratively until no free superpoints remain, yielding the $Q$ final 3D mask proposals $\{\M_q\}_{q=1}^{Q}$ and its score $\{o_q\}_{q=1}^{Q}$, where $Q = Q_1 + Q_2 + \ldots + Q_n$.

\begin{algorithm}[t]
\caption{Algorithm of 3D Mask Optimization.}
\label{alg:optimize}
\textbf{Input: }{Visibility vector $\overline{\M}^t_q$} in \cref{eq:visible_set}

\begin{algorithmic}[1]
\STATE {\bfseries initialize: } Objective value $\C_0 = \0$ and \\current solution $\theta_0 = \0$.

\FOR{$t = 1$ \TO $T$} 
    \STATE Option 1 - Retain current solution:\\
    $\theta_t^1 = \theta_{t-1}, \quad \C_t^1 = \C_{t-1}$
    
    \STATE Option 2 - Add all superpoints in $\overline{\M}^t_q$ to current solution: 
    $\theta_t^2 = \theta_{t-1} \vee \overline{\M}^t_q, \quad \C_t^2 = \mL(\theta_t)$ in \cref{eq:objective}

    \STATE {\bfseries If} $\C^1_t > \C^2_t$ 
    {\bfseries then:} 
    $\theta_t = \theta_t^1, \C_t = \C_t^1$ \\
    {\bfseries else:}
    $\theta_t = \theta_t^2, \C_t = \C_t^2$
    
\ENDFOR
\STATE {\bfseries return} $\theta_T$
\end{algorithmic}
\end{algorithm}

%% file: sec/4_experiments.tex
\section{Experiments}
\label{sec:experiment}

\subsection{Experimental Setup}

\myheading{Datasets:} To evaluate the effectiveness of our proposed method, we conduct experiments on two widely-used 3D instance segmentation datasets: ScanNet200 \cite{dai2017scannet} and ScanNet++ \cite{yeshwanthliu2023scannetpp}. ScanNet200 comprises approximately 1201 training scenes and 312 validation scenes, annotated with 200 object categories, facilitating detailed 3D semantic and instance segmentation. ScanNet++ offers posed RGB-D images and high-quality 3D geometry obtained via advanced laser scanning technology, including 360 training scenes and 50 validation scenes with annotations spanning up to 1,659 semantic classes. We evaluate our method on the validation sets across the three settings including class-agnostic, open-vocab, and open-ended 3DIS.

\myheading{Evaluation metrics:} We evaluate our method using the primary metric of $\mathbf{AP}$ over $\operatorname{IoU}$ thresholds from 50\% to 95\% in 5\% increments. Also, we report the metric at $\operatorname{IoU}$ thresholds of 50\% and 25\%, denoted as $\mathbf{AP_{50}}$ and $\mathbf{AP_{25}}$. For OV-3DIS settings on the ScanNet200 dataset, we further report category group-specific metrics: $\mathbf{AP_{head}}$, $\mathbf{AP_{com}}$, and $\mathbf{AP_{tail}}$, as commonly presented in prior work.

\myheading{Implementation details:} For 2D-3D image-point occlusion depth threshold, we use 0.1 for both datasets. For 3D lifting, we use $\tau=0.5$. We perform RGB-D sampling at an interval of 10 views, following the approach in \cite{yin2023sai3d, nguyen2023open3dis, takmaz2023openmask3d}. For class-agnostic 3D instance segmentation, we utilize the latest model of Segment Anything Model 2 (ViT-L) \cite{ravi2024sam}. For OV-3DIS setting, we employ the OpenAI CLIP ViT-L/14@336px model \cite{radford2021learning} as described in \cite{takmaz2023openmask3d} for pointwise feature extraction \cite{nguyen2023open3dis}, and for OE-3DIS, we leverage the OSM \cite{yu2023osm} for aggregating point cloud feature tokens \cite{nguyen2024open}, with Vicuna-7B as the LLM decoder.

\begin{table}[t!]
\small
\setlength{\tabcolsep}{6pt}
\centering
\begin{tabular}{lccccc}
\toprule
\textbf{Method} & \textbf{2D Segmenter} & \textbf{AP} &  \textbf{AP$_{50}$} & \textbf{AP$_{25}$}  \\
\midrule
Segment3D \cite{huang2025segment3d}& \multirow{6}{*}{SAM-HQ} & 19.0 & 29.7 & 41.6 \\
SAM-Graph \cite{guo2024sam-graph}& & 15.3 & 27.2 & 44.3 \\
SAM3D \cite{yang2023sam3d} &  & 8.3 & 17.5 & 33.7  \\
SAMPro3D \cite{xu2023sampro3d}&  & 16.9 & 31.7 & 48.6\\
SAI3D \cite{yin2023sai3d} &  & 17.1 & 31.1 & \textbf{49.5}\\
Open3DIS \cite{nguyen2023open3dis} &  & 20.7 & \textbf{38.6} & 47.1 \\

\midrule
Open3DIS \cite{nguyen2023open3dis} & \multirow{2}{*}{SAM2-L} & 17.9 & 30.4 & 39.7\\
\textbf{\Approach~(ours)} & & \textbf{22.2} & 35.8 & 47.0 \\

\bottomrule
\end{tabular}
\vspace{-4pt}
\caption{\textbf{Results of Class-Agnostic 3DIS on ScanNet++,} benchmarking on 1554 classes}
\label{tab:scannetpp_agnostic}
\vspace{-5pt}
\end{table}

\begin{table}[t!]
\small
\setlength{\tabcolsep}{6pt}
\centering
\begin{tabular}{lcccccc}
\toprule
\textbf{Method} & \textbf{2D Segmenter} & \textbf{AP} & \textbf{AP$_{50}$} & \textbf{AP$_{25}$}  \\
\midrule
OVIR-3D \cite{lu2023ovir}& \multirow{2}{*}{SAM-HQ} & 14.4 & 27.5 & 38.8 \\
Open3DIS \cite{nguyen2023open3dis} &  & 31.5 & 45.3 & 51.1\\
\midrule
Open3DIS \cite{nguyen2023open3dis} & \multirow{2}{*}{SAM2-L} & 26.7 & 40.2 & 53.8\\
\textbf{\Approach~(ours)} &  & \textbf{32.5} & \textbf{45.2} & \textbf{55.0} \\
\midrule
ISBNet \cite{ngo2023isbnet} & 3D & 40.2 & 50.0 & 54.6 \\
Open3DIS \cite{nguyen2023open3dis}& 3D + SAM-HQ & 41.5 & \textbf{51.6} & \textbf{56.3} \\
\textbf{\Approach~(ours)} & 3D + SAM2-L &  \textbf{42.5} & 51.2 & 54.5 \\

\bottomrule
\end{tabular}
\vspace{-4pt}
\caption{\textbf{Results of class-Agnostic 3DIS on ScanNet200,} benchmarking on 198 classes}
\label{tab:scannet200_agnostic}
\vspace{-10pt}
\end{table}

\subsection{Comparison to prior work}
\label{sec:comparison-prior-work}
\myheading{Setting 1: Class-Agnostic 3D Instance Segmentation.}
We begin by evaluating our method in the class-agnostic 3D instance segmentation setting, where the objective is to detect and segment instances without relying on predefined semantic categories. \cref{tab:scannetpp_agnostic} and \cref{tab:scannet200_agnostic} present the performance comparison between our approach and existing methods on the ScanNet++ and ScanNet200 datasets, respectively. 
For \textit{ScanNet++} (\cref{tab:scannet200_agnostic}), \Approach~ achieves an AP of 22.2, surpassing the previous state-of-the-art method Open3DIS \cite{nguyen2023open3dis}, which attained an AP of 17.9 (using the same segmenter, SAM2-L) and 20.7 (with the higher-quality segmenter SAM-HQ \cite{sam_hq}). This improvement underscores the effectiveness of our method in handling a wide variety of object categories. 
For \textit{ScanNet200} (\cref{tab:scannet200_agnostic}), our method achieves an AP of 32.5\% using SAM2-L, outperforming both OVIR-3D \cite{lu2023ovir} and Open3DIS \cite{nguyen2023open3dis}, which utilize SAM-HQ \cite{sam_hq} for generating 3D proposals from 2D masks alone. By integrating with 3D proposals from ISBNet \cite{ngo2023isbnet}, \Approach~ further boosts the AP to 42.5, surpassing existing methods. These results highlight the advantage of our approach over previous methods.

\begin{table}[t!]
\small
\setlength{\tabcolsep}{6pt}
\centering
\begin{tabular}{lcccccc}
\toprule
\textbf{Method} & \textbf{2D Segmenter} & \textbf{AP} & \textbf{AP$_{50}$} & \textbf{AP$_{25}$}  \\
\midrule
OVIR-3D \cite{lu2023ovir} & \multirow{3}{*}{SAM-HQ} & 3.6 & 5.7 & 7.3 \\
Segment3D \cite{huang2025segment3d} & & 10.1 & 17.7 & 20.2 \\
Open3DIS \cite{nguyen2023open3dis}  & & 11.9 & 18.1 & 21.7 \\
\midrule
\textbf{\Approach~(ours)} & SAM2-L & \textbf{12.9} & \textbf{19.0} & \textbf{21.9} \\

\bottomrule
\end{tabular}
\vspace{-4pt}
\caption{\textbf{Results of Open-Vocabulary 3DIS on ScanNet++}, benchmarking a subset of 100 classes.}
\label{tab:scannetpp_openvocab}
\vspace{-10pt}
\end{table}

\begin{table}[t!]
\small
\setlength{\tabcolsep}{4pt}
\centering
\begin{tabular}{lcccccc}
\toprule
\textbf{Method} & \textbf{Segmenter} & \textbf{AP} &  \textbf{AP$_{h}$} & \textbf{AP$_{c}$} & \textbf{AP$_{t}$}  \\
\midrule
Search3D \cite{takmaz2024search3d} & \multirow{2}{*}{Mask3D \cite{schult2023mask3d}} & 14.3 & 16.1 & 13.6 & 12.9 \\
OpenMask3D \cite{takmaz2023openmask3d} & & 15.4 &17.1 & 14.1 & 14.9\\
\midrule
SAI3D \cite{yin2023sai3d} & \multirow{2}{*}{SAM-HQ} & 12.7 & 12.1 & 10.4 & 16.2 \\
OVIR-3D \cite{lu2023ovir} &  & 11.2 & 12.6 & 11.5 & 10.3\\
\midrule
Open3DIS \cite{nguyen2023open3dis} &  SAM-HQ &18.2&\textbf{18.9}&16.5&19.2 \\
\textbf{\Approach~(ours)} & SAM2-L & \textbf{19.1}&17.5&\textbf{17.2}&\textbf{23.3}    \\
\midrule
Open3DIS \cite{nguyen2023open3dis} & 3D + SAM-HQ & 23.7&\textbf{27.8}&21.2&21.8  \\
\textbf{\Approach~(ours)} & 3D + SAM2-L &\textbf{25.8}& 27.4 & \textbf{23.8} & \textbf{26.4}   \\

\bottomrule
\end{tabular}
\vspace{-4pt}
\caption{\textbf{Results of Open-Vocabulary 3DIS on ScanNet200}, benchmarking on 198 classes.}
\label{tab:scannet200_openvocab}
\vspace{-10pt}
\end{table}

\myheading{Setting 2: Open-Vocabulary Instance Segmentation.}
We next evaluate our method in an open-vocabulary setting, where the model must segment instances from previously unseen categories. Following the challenge setup, we assess our approach on the 100-class subset of ScanNet++. On this dataset (\cref{tab:scannetpp_openvocab}), \Approach~achieves an AP of 12.9, outperforming Open3DIS\cite{nguyen2023open3dis}, which scored an AP of 11.9, demonstrating improved generalization to the challenging ScanNet++ dataset.
On the ScanNet200 dataset (\cref{tab:scannet200_openvocab}), our method achieves an AP of 25.8 when using 3D proposals combined with SAM2-L, surpassing Open3DIS and other baseline methods. Even with only 2D proposals, our approach reaches an AP of 19.1, exceeding previous methods under comparable conditions and underscoring our method’s strength without relying on class name information in 3D proposals. Moreover, our method maintains robust performance across different category frequency groups, as demonstrated by the $\text{AP}_{\text{head}}$ and $\text{AP}_{\text{com}}$ metrics. The $\text{AP}_{\text{tail}}$ also achieves notable results, indicating that our approach is effective in providing consistent and fine-grained segmentation

\myheading{Setting 3: Open-Ended Instance Segmentation.}
Finally, we assess our method in the open-ended setting \cite{nguyen2024open}, where the model must segment and recognize instances without predefined class labels during testing. \cref{tab:scannetpp_openended} and \cref{tab:scannet200_openended} present the performance of our approach alongside baseline methods proposed in \cite{nguyen2024open} on the ScanNet++ and ScanNet200 datasets, respectively.
On ScanNet++ (\cref{tab:scannetpp_openended}), our approach achieves an AP of 20.1, surpassing prior baselines like Point-Wise, which achieved an AP of 18.4. This improvement highlights our method’s capability to handle diverse object categories without a constrained vocabulary.
For ScanNet200 (\cref{tab:scannet200_openended}), our method achieves an AP of 19.1, outperforming Point-Wise by a margin of 3.1 AP. This demonstrates our approach’s effectiveness in the open-ended setting across category frequency groups.

\begin{table}
\small
\setlength{\tabcolsep}{6pt}
\centering
\begin{tabular}{lcccccc}
\toprule
\textbf{Method} & \textbf{Setting} & \textbf{AP} & \textbf{AP$_{50}$} & \textbf{AP$_{25}$}  \\
\midrule
Large-Vocab \cite{nguyen2024open} & 21K+ classes & 7.3 & 11.9 & 15.2 \\
Image-Tagging \cite{nguyen2024open} & RAM++ & 9.1 & 15.5 & 19.1\\
Mask-Wise \cite{nguyen2024open} & \multirow{2}{*}{OSM} & 16.3 & 24.8 & 29.0\\
Point-Wise \cite{nguyen2024open} &  & 18.4 & 29.4 & 33.6\\
\midrule
\textbf{\Approach~(ours)} & OSM & \textbf{20.1} & \textbf{30.4} & \textbf{35.5}  \\

\bottomrule
\end{tabular}
\vspace{-4pt}
\caption{\textbf{Results of Open-Ended 3DIS on ScanNet++}, benchmarking a subset of 100 classes.}
\label{tab:scannetpp_openended}
\vspace{-10pt}
\end{table}

\begin{table}
\small
\setlength{\tabcolsep}{4pt}
\centering
\begin{tabular}{lccccccc}
\toprule
\textbf{Method} & \textbf{Setting} & \textbf{AP}&\textbf{AP$_{h}$} & \textbf{AP$_{c}$} & \textbf{AP$_{t}$}  \\
\midrule
Large-Vocab \cite{nguyen2024open} & 21K+ classes & 8.5& 9.9 & 7.2 & 8.3\\
Image-Tagging \cite{nguyen2024open} & RAM++ &10.7& 11.6 & 11.0 & 9.3\\
Mask-Wise \cite{nguyen2024open} & \multirow{2}{*}{OSM} &14.4& 18.9 & 13.5 & 10.2\\
Point-Wise \cite{nguyen2024open} &  &16.0& \textbf{20.0} & 14.3 & 13.2\\
\midrule
\textbf{\Approach~(ours)} & OSM & \textbf{19.1} & 18.2 &\textbf{16.9} & \textbf{22.8} \\

\bottomrule
\end{tabular}
\vspace{-4pt}
\caption{\textbf{Results of Open-Ended 3DIS on ScanNet200}, benchmarking on 198 instance classes.}
\label{tab:scannet200_openended}
\vspace{-10pt}
\end{table}

\begin{figure*}[t]
  \centering
  \includegraphics[width=0.91\linewidth]{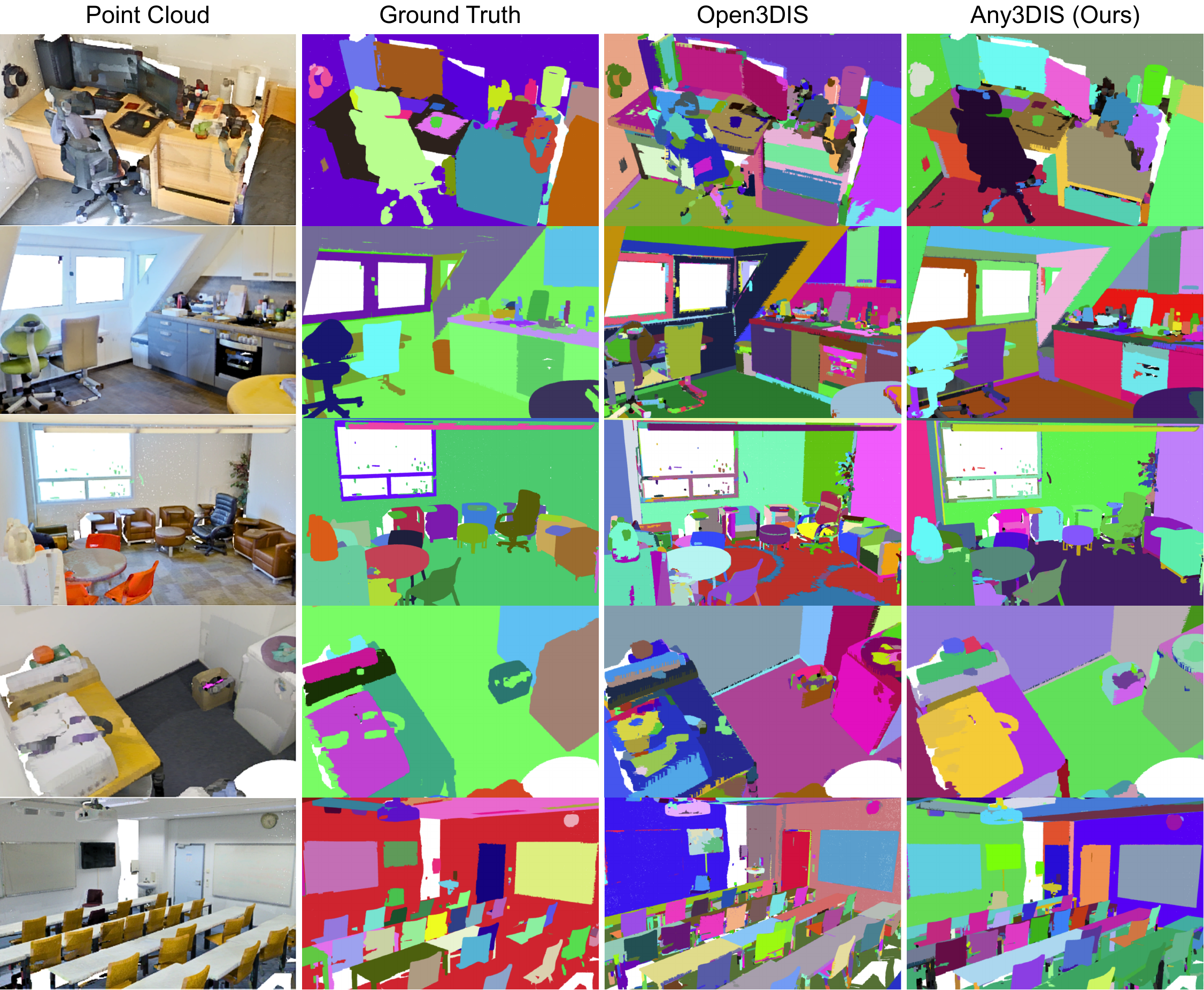}
   \vspace{-7pt}
   \caption{\textbf{Qualitative Result on ScanNet++ Validation Set.}: From left to right we show the input point cloud of 3D scenes, GT segmentation, Open3DIS \cite{nguyen2023open3dis}, and \Approach~(ours) results. Our approach achieves more accurate and consistent segmentation than Open3DIS.}
   \vspace{-12pt}
   \label{fig:qualitative-result}
\end{figure*}

\myheading{Qualitative results.}
In our qualitative results on the ScanNet++ validation set (\cref{fig:qualitative-result}), we compare the segmentation performance of \Approach~with Open3DIS and ground truth (GT). The leftmost column shows the input 3D point clouds, followed by GT masks, Open3DIS results, and our approach’s results. Unlike Open3DIS, which frequently exhibits over-segmentation and struggles to accurately group object instances, \Approach~produces more coherent segmentation, capturing object boundaries and structures with greater fidelity. Our method reduces redundancy and enhances segmentation accuracy across a variety of challenging indoor scenes.
These results underscore \Approach’s ability to achieve more precise and consistent segmentation through the 3D-Aware 2D Mask Tracking, which maintains continuity and precision across frames. This improvement is particularly evident in complex environments with multiple objects and diverse spatial arrangements, showing our method’s robustness in producing high-quality 3DIS.

\begin{table}[t]
\small
\setlength{\tabcolsep}{3pt}
\centering
\begin{tabular}{cccccccc}
\toprule

\textbf{3D-AMT} & \textbf{MaskOpt} & \textbf{AP} & \textbf{AP$_{50}$} & \textbf{AP}$_{\text{25}}$ & \textbf{RC} & \textbf{RC}$_{\text{50}}$ & \textbf{RC}$_{\text{25}}$ \\
\midrule
&   & 17.0 & 30.1 & 42.6 & 30 & 50.7 & 67.8 \\ 
\checkmark &   & 17.9 & 31.2 & 45.1 & 32.6 & 53.1 & 69.5  \\ 
&  \checkmark &  20.3 & 33.7 & 44.5 & 35.8 & 56.9 & 71.9 \\ 
\checkmark &  \checkmark & \textbf{22.2} &   \textbf{35.8} & \textbf{47.0} & \textbf{40.8} & \textbf{63.8} & \textbf{80.8}   \\ 

\bottomrule
\end{tabular}
\vspace{-5pt}
\caption{\textbf{Ablation on each contribution on class-agnostic 3D instance segmentation of ScanNet++}}
\vspace{-7pt}
\label{tab:ablation_compselect}
\end{table}

\begin{table}[t]
\small
\setlength{\tabcolsep}{5pt}
\centering
\begin{tabular}{clcccc}
\toprule

\textbf{No.}&\textbf{Strategy} & \textbf{AP} & \textbf{AP$_{50}$} & \textbf{AP}$_{\text{25}}$ \\
\midrule
 1 & all lifted superpoints & 17.9 & 31.2 & 45.1\\
 2 & superpoints in top-1 view & 19.8 & 31.4 & 42.9 \\
 3 & superpoints in top-5 views & 20.6 & 33.6 & 44.9 \\
 4 & superpoints in top-10 views & 20.8 & 33.5 & 44.4 \\
 5 & optimization (ours) & \textbf{22.2} & \textbf{35.8} & \textbf{47.0}\\
\bottomrule
\end{tabular}
\vspace{-5pt}
\caption{\textbf{Study on different strategies for Mask Optimization.}}
\vspace{-7pt}
\label{tab:opt_strategy}
\end{table}

\subsection{Ablation study}
In this section, we conduct ablation studies on ScanNet++ \cite{yeshwanthliu2023scannetpp} to analyze the impact of each component of our proposed method on the class-agnostic 3DIS performance.

\myheading{Study on the effect each component,} 3D-Aware 2D Mask Tracking (3D-AMT) and Mask Optimization (MaskOpt), is presented in \cref{tab:ablation_compselect}. The baseline method, which chooses the first frame as the pivot frame (excluding 3D-AMT) and utilizes all lifted superpoints without refinement (excluding MaskOpt), achieves an AP of 17.0. Incorporating 3D-AMT raises the AP to 17.9, highlighting that 3D-AMT enhances image feature representation for 3D segmentation. Adding MaskOpt alone significantly boosts the AP to 20.3, showcasing the impact of refining the mask selection process. Using both 3D-AMT and MaskOpt together results in the highest AP of 22.2, confirming that these components effectively complement each other to enhance performance. \textit{We provide a detailed analysis of each component and the total runtime in the supplementary material}!

\myheading{Mask optimization strategies.} We evaluate various strategies for mask optimization in \cref{tab:opt_strategy}. Using all lifted superpoints without any refinement (No.~1) achieves an AP of 17.9. In \cref{eq:objective}, rather than using all views in our proposed \cref{alg:optimize}, an alternative approach is to focus on the top views with the highest objective values, $\mL(\theta)$. We then exhaustively explore all possible configurations of superpoints visible in these selected views to determine the optimal configuration $\theta$. A top-1 view selection strategy (No.~2) slightly raises the AP to 19.8. Choosing the top-5 (No.~3) and top-10 (No.~4) views further improves APs to 20.6 and 20.8, respectively. Our optimization approach (No.~5) achieves the highest AP of 22.2, confirming its effectiveness in identifying a good combination of superpoints.

%% file: sec/5_conclusion.tex
\section{Limitations and Conclusion}
\label{sec:experiment}

\myheading{Limitations:} Despite its strengths, \Approach~has certain limitations. Its performance heavily relies on the quality of the 2D segmentation and tracking provided by SAM 2, meaning that inaccuracies in the 2D masks may propagate into the 3D segmentation results. Furthermore, the current 3D mask optimization process remains suboptimal due to the infeasibility of exhaustively exploring all possible configurations. This could be left for future studies. We anticipate that future improvements in SAM 2’s segmentation and tracking accuracy could enhance \Approach’s performance. Also, advancements in quantum computing could enable optimal solutions to the proposed unconstrained binary optimization problem.

\myheading{Conclusion:}
In this paper, we have introduced \Approach, a novel class-agnostic approach for 3D instance segmentation that leverages 2D mask tracking to segment 3D objects in point cloud scenes. By identifying the pivot view for each sampled superpoint and employing SAM 2 for 2D segmentation and tracking, our method effectively leverages the 2D foundation model for 3D instance segmentation tasks. The 3D proposal refinement through optimization further enhances the segmentation masks, ensuring high-quality 3D instance proposals. Extensive experiments on the ScanNet200 and ScanNet++ datasets demonstrate that \Approach~ consistently outperforms existing methods across various settings, including class-agnostic, open-vocabulary, and open-ended 3D instance segmentation, validating the effectiveness of our framework.

%% file: main.bbl
\begin{thebibliography}{36}
\providecommand{\natexlab}[1]{#1}
\providecommand{\url}[1]{\texttt{#1}}
\expandafter\ifx\csname urlstyle\endcsname\relax
  \providecommand{\doi}[1]{doi: #1}\else
  \providecommand{\doi}{doi: \begingroup \urlstyle{rm}\Url}\fi

\bibitem[Baruch et~al.(2021)Baruch, Chen, Dehghan, Dimry, Feigin, Fu, Gebauer, Joffe, Kurz, Schwartz, and Shulman]{dehghan2021arkitscenes}
Gilad Baruch, Zhuoyuan Chen, Afshin Dehghan, Tal Dimry, Yuri Feigin, Peter Fu, Thomas Gebauer, Brandon Joffe, Daniel Kurz, Arik Schwartz, and Elad Shulman.
\newblock {ARK}itscenes - a diverse real-world dataset for 3d indoor scene understanding using mobile {RGB}-d data.
\newblock In \emph{Thirty-fifth Conference on Neural Information Processing Systems Datasets and Benchmarks Track (Round 1)}, 2021.

\bibitem[Behley et~al.(2019)Behley, Garbade, Milioto, Quenzel, Behnke, Stachniss, and Gall]{behley2019iccv}
J. Behley, M. Garbade, A. Milioto, J. Quenzel, S. Behnke, C. Stachniss, and J. Gall.
\newblock {SemanticKITTI: A Dataset for Semantic Scene Understanding of LiDAR Sequences}.
\newblock In \emph{Proc. of the IEEE/CVF International Conf.~on Computer Vision (ICCV)}, 2019.

\bibitem[Cheng et~al.(2021)Cheng, Schwing, and Kirillov]{cheng2021maskformer}
Bowen Cheng, Alexander~G. Schwing, and Alexander Kirillov.
\newblock Per-pixel classification is not all you need for semantic segmentation.
\newblock In \emph{NeurIPS}, 2021.

\bibitem[Choy et~al.(2019)Choy, Gwak, and Savarese]{mink}
Christopher Choy, JunYoung Gwak, and Silvio Savarese.
\newblock 4d spatio-temporal convnets: Minkowski convolutional neural networks.
\newblock In \emph{Proceedings of the IEEE Conference on Computer Vision and Pattern Recognition}, pages 3075--3084, 2019.

\bibitem[Dai et~al.(2017)Dai, Chang, Savva, Halber, Funkhouser, and Nie{\ss}ner]{dai2017scannet}
Angela Dai, Angel~X. Chang, Manolis Savva, Maciej Halber, Thomas Funkhouser, and Matthias Nie{\ss}ner.
\newblock Scannet: Richly-annotated 3d reconstructions of indoor scenes.
\newblock In \emph{Proc. Computer Vision and Pattern Recognition (CVPR), IEEE}, 2017.

\bibitem[Deng(2020)]{dbscan}
Dingsheng Deng.
\newblock Dbscan clustering algorithm based on density.
\newblock In \emph{2020 7th International Forum on Electrical Engineering and Automation (IFEEA)}, pages 949--953, 2020.

\bibitem[Ding et~al.(2024)Ding, Yang, Xue, Zhang, Bai, and Qi]{ding2024lowis3d}
Runyu Ding, Jihan Yang, Chuhui Xue, Wenqing Zhang, Song Bai, and Xiaojuan Qi.
\newblock Lowis3d: Language-driven open-world instance-level 3d scene understanding.
\newblock \emph{IEEE Transactions on Pattern Analysis and Machine Intelligence}, 2024.

\bibitem[Eldar et~al.(1997)Eldar, Lindenbaum, Porat, and Zeevi]{eldar1997farthest}
Yuval Eldar, Michael Lindenbaum, Moshe Porat, and Yehoshua~Y Zeevi.
\newblock The farthest point strategy for progressive image sampling.
\newblock \emph{IEEE transactions on image processing}, 6\penalty0 (9):\penalty0 1305--1315, 1997.

\bibitem[Felzenszwalb and Huttenlocher(2004)]{felzenszwalb2004efficient}
Pedro~F Felzenszwalb and Daniel~P Huttenlocher.
\newblock Efficient graph-based image segmentation.
\newblock \emph{International journal of computer vision}, 59:\penalty0 167--181, 2004.

\bibitem[Guo et~al.(2024)Guo, Zhu, Peng, Wang, Shen, Hu, and Zhou]{guo2024sam-graph}
Haoyu Guo, He Zhu, Sida Peng, Yuang Wang, Yujun Shen, Ruizhen Hu, and Xiaowei Zhou.
\newblock Sam-guided graph cut for 3d instance segmentation.
\newblock In \emph{ECCV}, 2024.

\bibitem[He et~al.(2021)He, Shen, and van~den Hengel]{He2021dyco3d}
Tong He, Chunhua Shen, and Anton van~den Hengel.
\newblock {DyCo3d}: Robust instance segmentation of 3d point clouds through dynamic convolution.
\newblock In \emph{Proceedings of the IEEE Conference on Computer Vision and Pattern Recognition (CVPR)}, 2021.

\bibitem[Huang et~al.(2025)Huang, Peng, Takmaz, Tombari, Pollefeys, Song, Huang, and Engelmann]{huang2025segment3d}
Rui Huang, Songyou Peng, Ayca Takmaz, Federico Tombari, Marc Pollefeys, Shiji Song, Gao Huang, and Francis Engelmann.
\newblock Segment3d: Learning fine-grained class-agnostic 3d segmentation without manual labels.
\newblock In \emph{European Conference on Computer Vision}, pages 278--295. Springer, 2025.

\bibitem[Jiang et~al.(2020)Jiang, Zhao, Shi, Liu, Fu, and Jia]{jiang2020pointgroup}
Li Jiang, Hengshuang Zhao, Shaoshuai Shi, Shu Liu, Chi-Wing Fu, and Jiaya Jia.
\newblock Pointgroup: Dual-set point grouping for 3d instance segmentation.
\newblock \emph{Proceedings of the IEEE Conference on Computer Vision and Pattern Recognition (CVPR)}, 2020.

\bibitem[Ke et~al.(2023)Ke, Ye, Danelljan, Liu, Tai, Tang, and Yu]{sam_hq}
Lei Ke, Mingqiao Ye, Martin Danelljan, Yifan Liu, Yu-Wing Tai, Chi-Keung Tang, and Fisher Yu.
\newblock Segment anything in high quality.
\newblock In \emph{NeurIPS}, 2023.

\bibitem[Kirillov et~al.(2023)Kirillov, Mintun, Ravi, Mao, Rolland, Gustafson, Xiao, Whitehead, Berg, Lo, et~al.]{kirillov2023segment}
Alexander Kirillov, Eric Mintun, Nikhila Ravi, Hanzi Mao, Chloe Rolland, Laura Gustafson, Tete Xiao, Spencer Whitehead, Alexander~C Berg, Wan-Yen Lo, et~al.
\newblock Segment anything.
\newblock In \emph{Proceedings of the IEEE/CVF International Conference on Computer Vision}, pages 4015--4026, 2023.

\bibitem[Lu et~al.(2023{\natexlab{a}})Lu, Kuen, Tiancheng, Jiuxiang, Weidong, Jiaya, Zhe, and Ming-Hsuan]{qilu2023high}
Qi Lu, Jason Kuen, Shen Tiancheng, Gu Jiuxiang, Guo Weidong, Jia Jiaya, Lin Zhe, and Yang Ming-Hsuan.
\newblock High-quality entity segmentation.
\newblock In \emph{ICCV}, 2023{\natexlab{a}}.

\bibitem[Lu et~al.(2023{\natexlab{b}})Lu, Chang, Jing, Boularias, and Bekris]{lu2023ovir}
Shiyang Lu, Haonan Chang, Eric~Pu Jing, Abdeslam Boularias, and Kostas Bekris.
\newblock Ovir-3d: Open-vocabulary 3d instance retrieval without training on 3d data.
\newblock In \emph{Conference on Robot Learning}, pages 1610--1620. PMLR, 2023{\natexlab{b}}.

\bibitem[Nguyen et~al.(2024{\natexlab{a}})Nguyen, Luu, Tran, Pham, and Nguyen]{nguyen2024open}
Phuc~DA Nguyen, Minh Luu, Anh Tran, Cuong Pham, and Khoi Nguyen.
\newblock Open-ended 3d point cloud instance segmentation.
\newblock \emph{arXiv preprint arXiv:2408.11747}, 2024{\natexlab{a}}.

\bibitem[Nguyen et~al.(2024{\natexlab{b}})Nguyen, Ngo, Kalogerakis, Gan, Tran, Pham, and Nguyen]{nguyen2023open3dis}
Phuc D.~A. Nguyen, Tuan~Duc Ngo, Evangelos Kalogerakis, Chuang Gan, Anh Tran, Cuong Pham, and Khoi Nguyen.
\newblock Open3dis: Open-vocabulary 3d instance segmentation with 2d mask guidance.
\newblock In \emph{Proceedings of the IEEE/CVF Conference on Computer Vision and Pattern Recognition (CVPR)}, 2024{\natexlab{b}}.

\bibitem[Qi et~al.(2017{\natexlab{a}})Qi, Su, Mo, and Guibas]{qi2017pointnet}
Charles~R Qi, Hao Su, Kaichun Mo, and Leonidas~J Guibas.
\newblock Pointnet: Deep learning on point sets for 3d classification and segmentation.
\newblock In \emph{Proceedings of the IEEE conference on computer vision and pattern recognition}, pages 652--660, 2017{\natexlab{a}}.

\bibitem[Qi et~al.(2017{\natexlab{b}})Qi, Yi, Su, and Guibas]{qi2017pointnet++}
Charles~Ruizhongtai Qi, Li Yi, Hao Su, and Leonidas~J Guibas.
\newblock Pointnet++: Deep hierarchical feature learning on point sets in a metric space.
\newblock \emph{Advances in neural information processing systems}, 30, 2017{\natexlab{b}}.

\bibitem[Radford et~al.(2021)Radford, Kim, Hallacy, Ramesh, Goh, Agarwal, Sastry, Askell, Mishkin, Clark, et~al.]{radford2021learning}
Alec Radford, Jong~Wook Kim, Chris Hallacy, Aditya Ramesh, Gabriel Goh, Sandhini Agarwal, Girish Sastry, Amanda Askell, Pamela Mishkin, Jack Clark, et~al.
\newblock Learning transferable visual models from natural language supervision.
\newblock In \emph{International conference on machine learning}, pages 8748--8763. PMLR, 2021.

\bibitem[Ravi et~al.(2024)Ravi, Gabeur, Hu, Hu, Ryali, Ma, Khedr, R{\"a}dle, Rolland, Gustafson, et~al.]{ravi2024sam}
Nikhila Ravi, Valentin Gabeur, Yuan-Ting Hu, Ronghang Hu, Chaitanya Ryali, Tengyu Ma, Haitham Khedr, Roman R{\"a}dle, Chloe Rolland, Laura Gustafson, et~al.
\newblock Sam 2: Segment anything in images and videos.
\newblock \emph{arXiv preprint arXiv:2408.00714}, 2024.

\bibitem[Rozenberszki et~al.(2022)Rozenberszki, Litany, and Dai]{rozenberszki2022language}
David Rozenberszki, Or Litany, and Angela Dai.
\newblock Language-grounded indoor 3d semantic segmentation in the wild.
\newblock In \emph{Proceedings of the European Conference on Computer Vision ({ECCV})}, 2022.

\bibitem[Schult et~al.(2023)Schult, Engelmann, Hermans, Litany, Tang, and Leibe]{schult2023mask3d}
Jonas Schult, Francis Engelmann, Alexander Hermans, Or Litany, Siyu Tang, and Bastian Leibe.
\newblock Mask3d: Mask transformer for 3d semantic instance segmentation.
\newblock In \emph{2023 IEEE International Conference on Robotics and Automation (ICRA)}, pages 8216--8223. IEEE, 2023.

\bibitem[Straub et~al.(2019)Straub, Whelan, Ma, Chen, Wijmans, Green, Engel, Mur-Artal, Ren, Verma, et~al.]{straub2019replica}
Julian Straub, Thomas Whelan, Lingni Ma, Yufan Chen, Erik Wijmans, Simon Green, Jakob~J Engel, Raul Mur-Artal, Carl Ren, Shobhit Verma, et~al.
\newblock The replica dataset: A digital replica of indoor spaces.
\newblock \emph{arXiv preprint arXiv:1906.05797}, 2019.

\bibitem[Takmaz et~al.(2023)Takmaz, Fedele, Sumner, Pollefeys, Tombari, and Engelmann]{takmaz2023openmask3d}
Ay{\c{c}}a Takmaz, Elisabetta Fedele, Robert~W. Sumner, Marc Pollefeys, Federico Tombari, and Francis Engelmann.
\newblock {OpenMask3D: Open-Vocabulary 3D Instance Segmentation}.
\newblock In \emph{Advances in Neural Information Processing Systems (NeurIPS)}, 2023.

\bibitem[Takmaz et~al.(2024)Takmaz, Delitzas, Sumner, Engelmann, Wald, and Tombari]{takmaz2024search3d}
Ayca Takmaz, Alexandros Delitzas, Robert~W Sumner, Francis Engelmann, Johanna Wald, and Federico Tombari.
\newblock Search3d: Hierarchical open-vocabulary 3d segmentation.
\newblock \emph{arXiv preprint arXiv:2409.18431}, 2024.

\bibitem[Tuan Duc~Ngo(2023)]{ngo2023isbnet}
Khoi~Nguyen Tuan Duc~Ngo, Binh-Son~Hua.
\newblock Isbnet: a 3d point cloud instance segmentation network with instance-aware sampling and box-aware dynamic convolution.
\newblock In \emph{Proceedings of the IEEE/CVF Conference on Computer Vision and Pattern Recognition (CVPR)}, 2023.

\bibitem[Vu et~al.(2022)Vu, Kim, Luu, Nguyen, and Yoo]{vu2022softgroup}
Thang Vu, Kookhoi Kim, Tung~M. Luu, Xuan~Thanh Nguyen, and Chang~D. Yoo.
\newblock Softgroup for 3d instance segmentation on 3d point clouds.
\newblock In \emph{CVPR}, 2022.

\bibitem[Xu et~al.(2023)Xu, Yin, Qiu, Liu, Tong, and Han]{xu2023sampro3d}
Mutian Xu, Xingyilang Yin, Lingteng Qiu, Yang Liu, Xin Tong, and Xiaoguang Han.
\newblock Sampro3d: Locating sam prompts in 3d for zero-shot scene segmentation.
\newblock \emph{arXiv preprint arXiv:2311.17707}, 2023.

\bibitem[Yan et~al.(2024)Yan, Zhang, Zhu, and Wang]{yan2024maskclustering}
Mi Yan, Jiazhao Zhang, Yan Zhu, and He Wang.
\newblock Maskclustering: View consensus based mask graph clustering for open-vocabulary 3d instance segmentation.
\newblock In \emph{Proceedings of the IEEE/CVF Conference on Computer Vision and Pattern Recognition}, pages 28274--28284, 2024.

\bibitem[Yang et~al.(2023)Yang, Wu, He, Zhao, and Liu]{yang2023sam3d}
Yunhan Yang, Xiaoyang Wu, Tong He, Hengshuang Zhao, and Xihui Liu.
\newblock Sam3d: Segment anything in 3d scenes.
\newblock \emph{arXiv preprint arXiv:2306.03908}, 2023.

\bibitem[Yeshwanth et~al.(2023)Yeshwanth, Liu, Nie{\ss}ner, and Dai]{yeshwanthliu2023scannetpp}
Chandan Yeshwanth, Yueh-Cheng Liu, Matthias Nie{\ss}ner, and Angela Dai.
\newblock Scannet++: A high-fidelity dataset of 3d indoor scenes.
\newblock In \emph{Proceedings of the International Conference on Computer Vision ({ICCV})}, 2023.

\bibitem[Yin et~al.(2024)Yin, Liu, Xiao, Cohen-Or, Huang, and Chen]{yin2023sai3d}
Yingda Yin, Yuzheng Liu, Yang Xiao, Daniel Cohen-Or, Jingwei Huang, and Baoquan Chen.
\newblock Sai3d: Segment any instance in 3d scenes.
\newblock In \emph{Proceedings of the IEEE/CVF Conference on Computer Vision and Pattern Recognition (CVPR)}, 2024.

\bibitem[Yu et~al.(2024)Yu, Shen, and Chen]{yu2023osm}
Qihang Yu, Xiaohui Shen, and Liang-Chieh Chen.
\newblock Towards open-ended visual recognition with large language model.
\newblock In \emph{ECCV}, 2024.

\end{thebibliography}
